\documentclass{nature}
\usepackage{hyperref}
\usepackage{cellspace}
\usepackage[flushleft]{threeparttable}
\usepackage{graphicx}
\usepackage{epstopdf}
\usepackage{dcolumn}
\usepackage{amssymb}
\usepackage{amsmath}
\usepackage{bm}

\title{A generalized method toward drug-target interaction prediction via low-rank matrix projection}

\author{Ratha Pech$^{1,2}$, Dong Hao$^{1,2}$, Yan-Li Lee$^{1,2}$, Maryna Po$^{3}$, Tao Zhou$^{1,2}$}

\begin{document}

\maketitle

\begin{affiliations}
 \item CompleX Lab, University of Electronic Science and Technology of China, Chengdu 611731, People's Republic of China.
 \item Big Data Research Center, University of Electronic Science and Technology of China, Chengdu 611731, People's Republic of China.
 \item Department of Chemistry and Biochemistry, George Mason University, Virginia, 22030, USA
\end{affiliations}

\begin{abstract}
Drug-target interaction (DTI) prediction plays a very important role in drug development and drug discovery. Biochemical experiments or \textit{in vitro} methods are very expensive, laborious and time-consuming. Therefore, \textit{in silico} approaches including docking simulation and machine learning have been proposed to solve this problem. In particular, machine learning approaches have attracted increasing attentions recently. However, in addition to the known drug-target interactions, most of the machine learning methods require extra characteristic information such as chemical structures, genome sequences, binding types and so on. Whenever such information is not available, they may perform poor. Very recently, the similarity-based link prediction methods were extended to bipartite networks, which can be applied to solve the DTI prediction problem by using topological information only. In this work, we propose a method based on low-rank matrix projection to solve the DTI prediction problem. On one hand, when there is no extra characteristic information of drugs or targets, the proposed method utilizes only the known interactions. On the other hand, the proposed method can also utilize the extra characteristic information when it is available and the performances will be remarkably improved. Moreover, the proposed method can predict the interactions associated with new drugs or targets of which we know nothing about their associated interactions, but only some characteristic information. We compare the proposed method with ten baseline methods, e.g., six similarity-based methods that utilize only the known interactions and four methods that utilize the extra characteristic information.
\end{abstract}

\section{Introduction}
Developing a new drug to the market is very costly and usually takes too much time \cite{dimasi2001new,chen2015drug}. Therefore, in order to save time and cost, scientists have tried to identify new uses of existing drugs, known as drug repositioning. Moreover, predicting these interactions between drugs and targets is one of the most active domains in drug research since they can help in the drug discovery \cite{hopkins2009drug,wishart2006drugbank,duran2017pioneering}, drug side-effect \cite{lounkine2012large,pauwels2011predicting} and drug repositioning \cite{cheng2012prediction,dudley2011exploiting,swamidass2011mining,moriaud2011identify}. Currently, the known interactions between drugs and target proteins are very limited \cite{dobson2004chemical,kanehisa2006genomics}, while it is believed that any single drug can interact with multiple targets \cite{chong2007new,macdonald2006identifying,xie2012novel,nascimento2016multiple}. However, laboratory experiments of biochemical verification on drug-target interactions are extremely expensive, laborious and time-consuming \cite{whitebread2005keynote,haggarty2003multidimensional,kuruvilla2002dissecting} since there are too many possible interactions to check. Thanks to the increasing capability to collect, store and process large-scale chemical and protein data, \textit{in silico} prediction of the interactions between drugs and target proteins becomes an effective and efficient tool for discovering the new uses of existing drugs. The prediction results provide helpful evidences to select potential candidates of drug-target interactions for further biochemical verification, which can reduce the costs and risks of failed drug development \cite{liu2016neighborhood,ashburn2004drug}.

There are two major approaches in \textit{in silico} prediction, including docking simulation and machine learning \cite{ding2013similarity}. Docking simulation is a common method in biology, but it has two major limitations. Firstly, this method requires three-dimensional structures of targets to compute the binding of each drug candidate \cite{halperin2002principles,rarey1996fast,shoichet1992molecular}, but such kind of information is usually not available \cite{ballesteros2001g,klabunde2002drug}. Secondly, it is very time-consuming. Therefore, in the last decade, many efforts have been made to solve the DTI prediction problem by machine learning approaches \cite{ding2013similarity,yamanishi2008prediction,bleakley2009supervised}. Most known machine learning methods \cite{nagamine2007statistical,nagamine2009integrating,yabuuchi2011analysis} treat DTI prediction as a binary classification prediction in which the drug-target interactions are regarded as instances and the characteristics of the drugs and target proteins are considered as features. To train the classifier, machine learning methods require label data, e.g., positive samples of truly existing interactions and negative samples of noninteractive pairs. Normally, the positive samples are available, but the negative samples are not known.

Whenever extra characteristic information about the drugs and targets are not available and only a portion of interactions between drugs and targets are known, similarity-based methods are suitable to solve this problem. However, there are also some limitations of the similarity-based methods. Firstly, similarity-based methods are not applicable for new drugs or targets that do not have any interactions at all since the methods cannot compute their similarities with the others. Secondly, although similarity-based methods are simple, sometimes they do not perform well \cite{pech2017link} since common neighbor and Jaccard indices, Cannistraci resource allocation (CRA) \cite{daminelli2015common} and Cannistraci Jaccard index (CJC) \cite{daminelli2015common} utilize only local information of networks. Katz index also performs unsatisfactorily (will show later) since the large decay factor will bring redundant information \cite{zhou2009accurate}, while the small decay factor makes Katz index close to common neighbor index or local path index \cite{zhou2009predicting,lu2009similarity}.

To overcome the limitations of machine learning methods and similarity-based methods, we propose a matrix-based method, namely low-rank matrix projection (LMP). LMP does not require negative samples. When the extra characteristic information of drugs and target proteins are not available, LMP utilizes only the known interactions. On the other hand, if the extra information is available, LMP can also take such information into consideration and remarkably improve the performances. LMP has been shown to perform better than similarity-based methods on the five renown datasets (e.g., MATADOR, enzyme, ion channel, GPCR and nuclear receptor). By embedding extra information of drugs and targets, LMP has been shown to outperform many baseline methods that also use extra information. Finally, LMP can effectively predict the potential interactions of new drugs or targets that do not have any known interactions at all. The proposed method can help selecting the most likely existing interactions for further chemical verifications in case only interaction information is known or only some of characteristic information is available. In a word, LMP can reduce cost and failure in drug development, and thus advance drug discoveries.


\begin{figure*}[ht!]
\centering
	\includegraphics[width=0.95\textwidth]{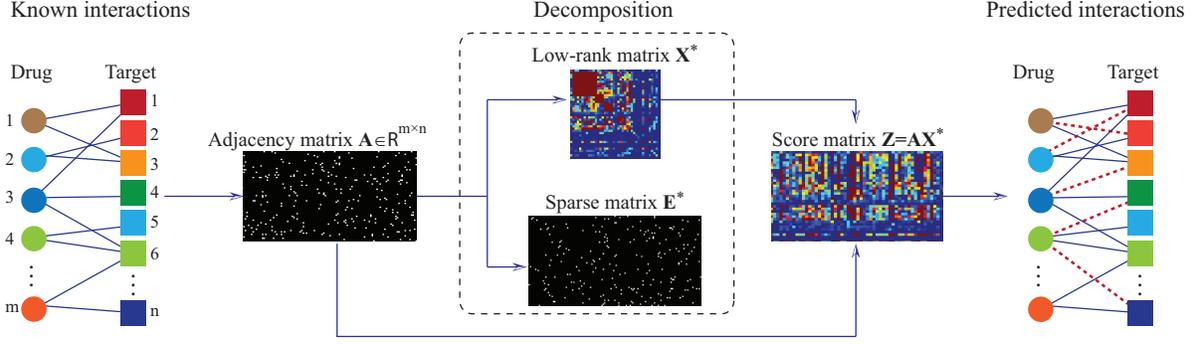}
    \caption{The illustration of the proposed method. Firstly, the known drug-target interactions are utilized to construct the adjacency matrix $\bf A$. Secondly, $\bf A$ is decomposed into a low-rank matrix ${\bf X}^*$ and a sparse matrix ${\bf E}^*$, which can be used to depict the hidden pattern and the noise in the original data. Finally, the score matrix is computed by projecting the adjacency matrix onto a lower-dimensional space via the low-rank matrix ${\bf X}^*$.}
\label{fig_framework}
\end{figure*}

\section{Methods}
\subsection{Notations}
In this work, various matrices and their similarity matrices are computed by the proposed method based on the characteristic information of the biological data. We denote them as follows:

\begin{itemize}
    \item $\mathbf{A}$: the adjacency matrix denoted the known interactions between drugs and proteins and its entries are defined as in Eq. (\ref{eq_adj_mat}).
    \item $\mathbf{X}_{D}^*$: the low-rank similarity matrix computed based on drug information in $\mathbf{A}$, i.e., similar drugs interact with similar targets.
    \item $\mathbf{X}_{T}^*$: the low-rank similarity matrix computed based on target information of $\mathbf{A}^T$ (transpose of $\mathbf{A}$), i.e., similar targets is interacted by similar drugs.
    \item $\mathbf{Z}_{AD}$: the score matrix obtained from the projection of $\mathbf{A}$ onto $\mathbf{X}_{D}^*$.
    \item $\mathbf{Z}_{AT}$: the score matrix obtained from the projection of $\mathbf{A}$ onto $\mathbf{X}_{T}^*$.
    \item $\mathbf{Z}_A$: the score matrix computed by combining $\mathbf{Z}_{AD}$ and $\mathbf{Z}_{AT}$.
    \item $\mathbf{S}_D$: the similarity matrix of drug compound computed by SIMCOMP \cite{hattori2003development} from the chemical structures of drugs which are obtained from KEGG LIGAND \cite{kanehisa2006genomics}.
    \item $\mathbf{X}_{SD}^*$: the low-rank similarity matrix computed by the proposed method on $\mathbf{S}_D$.
    \item $\mathbf{Z}_{D}$: the score matrix obtained by projecting $\mathbf{A}$ on $\mathbf{X}_{SD}^*$.
    \item $\mathbf{S}_T$: the similarity matrix of protein sequences computed by a normalized Smith-Waterman score \cite{smith1981identification} from GEGG GENES \cite{kanehisa2006genomics}.
    \item $\mathbf{X}_{ST}^*$: the low-rank similarity matrix computed by the proposed method on $\mathbf{S}_T$.
    \item $\mathbf{Z}_{T}$: the score matrix obtained by projecting $\mathbf{A}$ on $\mathbf{X}_{ST}^*$.
    \item $\mathbf{Z}_{ADT}$: the score matrix computed by combining $\mathbf{Z}_{A}$, $\mathbf{Z}_{D}$ and $\mathbf{Z}_{T}$.
\end{itemize}

\subsection{Datasets}
In this work, we implement the proposed method as well as similarity-based methods on five benchmark and renown datasets namely MATADOR \cite{gunther2007supertarget}, enzyme \cite{yamanishi2008prediction}, ion channel \cite{yamanishi2008prediction}, G-protein-coupled receptors (GPCR) \cite{yamanishi2008prediction}, and nuclear receptors \cite{yamanishi2008prediction}. The manually annotated target and drug online resource (MATADOR) (May 2017) dataset is a  free online dataset of chemical and target protein interactions. There are 13 columns in the dataset, however, we utilize only two columns, including Chemical ID and Protein ID, to construct the adjacency matrix. MATADOR dataset has no characteristic information about the drugs and targets. Enzyme, ion channel, GPCR, and nuclear receptors (May 2017) are the drug-target interaction networks for human beings. The statistics of the five datasets are presented in table \ref{table_stat}.

\begin{table}
\caption{ The statistics of the five datasets.}
    \begin{tabular}{p{3.5cm} p{2.7cm} p{2.7cm} p{2.5cm} p{2.5cm}}
    \hline \\ [-5.5ex]
              Dataset   &   Drug & Target  & Interaction & Sparsity of $\mathbf{A}$   \\
               \hline \\ [-5.5ex]
              MATADOR           & 801   & 2901  & 15843 & 0.007     \\
              enzyme            & 445   & 664   & 2926  & 0.010     \\
              ion channel       & 210   & 204   & 1476  & 0.034     \\
              GPCR              & 223   & 95    & 635   & 0.030     \\
              nuclear receptors & 54    & 26    & 90    & 0.061     \\  [0.5ex]
        \hline \\
      \end{tabular}
      \label{table_stat}
\end{table}

\subsection{Low-Rank Matrix Projection}
In the real-world problems, many data that are lying on the high-dimensional space and full of noise normally contain hidden features which can be seen after they are projected onto the lower dimensional space and simultaneously the noise are subtracted from them \cite{vidal2011tutorial}. Low-rank matrix has been shown to be a powerful and suitable tool to capture the patterns in high dimensional-space and noisy data \cite{wright2009robust,lin2010augmented,peng2012rasl}. Therefore, it is deserved to be investigated to solve DTI problem. In this section we assume that only known interactions between drugs and targets are available so we aim at learning the low-rank matrix from this interaction information. First of all, we construct the adjacency matrices of the drug-target interactions in the five datasets. Mathematically, the adjacency matrix is defined as
\begin{equation}
    \mathbf{A}_{ij}= \left\{ \begin{array}{l l l}
                             1, \ \ \ \textrm{if drug } i \textrm{ interacts with target } j \\
                             0, \ \ \ \textrm{otherwise}   \end{array}\right. .
\label{eq_adj_mat}
\end{equation}
Then we obtain $\mathbf{A}\in\mathbb{R}^{m\times n}$, where $m$ is the number of drugs and $n$ is the number of target proteins.

The real data are normally far from perfect, meaning that a portion of the drug-target interactions in the real data may be incorrect or redundant, and also some other drug-target interactions may be missing from the observed data. Therefore, the adjacency matrix $\mathbf{A}$ can be decomposed into two parts. The first part is a linear combination of $\mathbf{A}$ with the low-rank matrix, which is essentially a projection from the noisy data $\bf A$ into a more refined or informative and lower-dimensional space. The second part can be considered as the noise or the outliers, which is strained off from the original data $\bf A$ and represented by a sparse matrix with most entries being zeros. The method seeks the lowest-rank matrix among all the candidates which is further utilized to construct the score matrix that estimates the likelihoods of the potential interactions.

\begin{table}
\begin{threeparttable}
\centering
\small{
\caption{The illustration of the Inexact ALM algorithm}
\begin{tabular}{p{1.1cm} p{1.3cm} p{10.0cm}}
\hline \\ [-5.5ex]
\multicolumn{3}{l}{\textbf{Algorithm 1}: Solving problem of Eq. (\ref{eq_unconstraint_lrr}) by Inexact ALM } \\
\hline \\ [-5.5ex]
\multicolumn{3}{l}{\textbf{Input:} Given a dataset $\mathbf{A}$ parameters $\alpha$} \\
\multicolumn{3}{l}{\textbf{Output:} $\mathbf{X}^*$ and $\mathbf{E}^*$} \\ [0.8ex]
\multicolumn{3}{l}{\textbf{Initialize:} $\mathbf{X}=0, \mathbf{E}=0, \mathbf{Y}_1=0, \mathbf{Y}_2=0, \mu=10^{-4}$, $\max_{\mu}=10^{10}, \rho=1.1, \epsilon=10^{-8}$} \\
\multicolumn{3}{l}{\textbf{while} not converged do}\\
& 1. & fix the other and update $\mathbf{J}$ by \\
& & $\mathbf{J} = \arg \min \frac{1}{\mu}||\mathbf{J}||_* +\frac{1}{2}||\mathbf{J-(X+Y}_2/\mu)||_F^2$ \\
& 2. & fix the other and update $\mathbf{X}$ by \\
& & \multicolumn{1}{l}{ \ $\mathbf{X} = (\mathbf{I}+\mathbf{A}^T\mathbf{A})^{-1}$ $\left(\mathbf{A}^T\mathbf{A}-\mathbf{A}^T\mathbf{E}+\mathbf{J}+(\mathbf{A}^T\mathbf{Y}_1-\mathbf{Y}_2)/\mu\right)$} \\
& 3. & fix the other and update $\mathbf{E}$ by\\
& \multicolumn{2}{l}{$\mathbf{E} = \arg \min \frac{\alpha}{\mu}||\mathbf{E}||_{2,1} + \frac{1}{2}||\mathbf{E-(A-AX+Y}_1/\mu)||_F^2$}\\
& 4. & update the multiplier\\
& & $\mathbf{Y}_1 = \mathbf{Y}_1 + \mu(\mathbf{A-AX-E})$ \\
& & $\mathbf{Y}_2 = \mathbf{Y}_2 + \mu(\mathbf{X-J})$ \\ 
& 5. & update parameter $\mu$ by \\
& & $\mu = min(\rho \mu,\max_{\mu})$  \\
& 6. & check the convergence condition \\
& & \ \ \ $||\mathbf{A-AX-E}||_{\infty} < \epsilon $ and $||\mathbf{X-J}||_{\infty} < \epsilon $\\
\multicolumn{3}{l}{\textbf{end while}}\\[0.3ex]
\hline
\end{tabular}
}
\label{table:algoLRR}
\begin{tablenotes}
      \small
      \item \small{The setting of the hyperparameters follows the implementation of \cite{lin2010augmented}:
          } Since as stated in the literature, they are the optimal ones.
    \end{tablenotes}
  \end{threeparttable}
\end{table}

Firstly, we decompose $\mathbf{A}$ as follows,
\begin{equation}
    \mathbf{A=AX+E}.
\label{eq_decom}
\end{equation}
Obviously, there are infinite solutions of Eq. (\ref{eq_decom}). However, since we wish $\mathbf{X}$ to be low-rank, where rank of a matrix is the maximum number of linearly independent column (or row) vectors in the matrix, and $\mathbf{E}$ to be sparse, we can enforce the nuclear norm or trace norm on $\mathbf{X}$ and sparse norm on $\mathbf{E}$. Mathematically, Eq. (\ref{eq_decom}) can be thus relaxed as
\begin{equation}
    \min_{\mathbf{X,E}} ||\mathbf{X}||_*+\alpha||\mathbf{E}||_{2,1} \ \ \ \mathrm{s.t.} \ \ \ \mathbf{A = AX+E},
\label{eq_lrr}
\end{equation}
where $||\mathbf{A}||_*=\sum_i{\sigma_i}$ (i.e., $\sigma_i$ is the singular values of $\mathbf{A}$), $||\mathbf{E}||_{2,1}=\sum_{j=1}^n\sqrt{\sum_{i=1}^m(\mathbf{E}_{ij})^2}$ is the noise regularization strategy and $\alpha$ is a positive free parameter taking a role to balance the weights of low-rank matrix and sparse matrix. Minimizing the trace norm of a matrix well favors the lower-rank matrix, meanwhile the sparse norm is capable of identifying noise and outliers.

Eq. (\ref{eq_lrr}) can also be regarded as a generalization of the robust PCA \cite{liu2013robust,liu2010robust} because if the matrix $\mathbf{A}$ in $\mathbf{AX}$ in the right side of Eq. (\ref{eq_lrr}) is set as identity matrix, then the model is degenerated to the robust PCA. Eq. (\ref{eq_lrr}) can be rewritten into an equivalent problem as,
\begin{equation}
    \min_{\mathbf{X,E,J}}||\mathbf{J}||_*+\alpha||\mathbf{E}||_{2,1} \ \ \ \mathrm{s.t.} \ \ \ \mathbf{A = AX + E, X=J}.
    \label{eq_alm_equivalent}
\end{equation}
Eq. (\ref{eq_alm_equivalent}) is the constraint and convex optimization problem which can be solved by many off-the-self methods, e.g., iterative method (IT) \cite{cai2010singular}, accelerated proximal gradient (APG) \cite{wright2009robust}, dual approach \cite{lin2009fast}, and augmented Lagrange multiplier (ALM) \cite{lin2010augmented}. In this work, we employ Inexact ALM method by firstly converting Eq. (\ref{eq_alm_equivalent}) to an unconstraint problem, then minimize this problem by utilizing augmented Lagrange function such that

\begin{equation}
\begin{split}
    &\mathcal{L}= ||\mathbf{J}||_*+\alpha||\mathbf{E}||_{2,1}+\mathrm{tr} \left(\mathbf{Y}_1^T\left(\mathbf{A-AX-E}\right)\right) + \\
    &\mathrm{tr}\left(\mathbf{Y}_2^T\left(\mathbf{X-J}\right)\right)+\frac{\mu}{2}\left(||\mathbf{A-AX-E}||_F^2+||\mathbf{X-J}||_F^2\right),
\end{split}
\label{eq_unconstraint_lrr}
\end{equation}
where $\mu \ge 0$ is a penalty parameter and $\mathrm{tr(.)}$ is the trace norm. The Eq. (\ref{eq_unconstraint_lrr}) is unconstraint and can be solved by minimizing with respect to $\mathbf{J,X}$ and $\mathbf{E}$, respectively, by fixing the other variables and then updating the Lagrange multipliers $\mathbf{Y}_1,\mathbf{Y}_2$.
The detailed illustration of how to solve Eq. (\ref{eq_unconstraint_lrr}) is shown in table \ref{table:algoLRR}.

We denote the solution of Eq. (\ref{eq_unconstraint_lrr}) as $\mathbf{X}_T^*$, if $\mathbf{A}_{ij}$ represents the interaction drug $i$ and protein $j$, then $\mathbf{X}_T^* \in \mathbb{R}^{n\times n}$. It can be considered as a similarity matrix that describes similarity between proteins. While if $\mathbf{A}_{ij}$ represents the interactions between protein $i$ and drug $j$ (as the transposition of the adjacency matrix in Eq. (\ref{eq_decom})), then the solution of Eq. (\ref{eq_unconstraint_lrr}) is denoted as $\mathbf{X}_D^*\in\mathbb{R}^{m\times m}$ which describes similarity between drugs. After obtaining these two similarity matrices, we project the adjacency matrix $\mathbf{A}$ onto these lower dimensional spaces, respectively, as
\begin{equation}
    \mathbf{Z}_{AD}=\mathbf{A}^T\mathbf{X}^*_D \ \ \textrm{and} \ \ \mathbf{Z}_{AT}=\mathbf{A}\mathbf{X}^*_T.
    \label{eq_similarity}
\end{equation}
Finally, we combine the two similarity matrices as
\begin{equation}
    \mathbf{Z}_A=\frac{\mathbf{Z}^T_{AD}+{\mathbf{Z}_{AT}}}{2}.
    \label{eq_similarity_com}
\end{equation}

After obtaining $\mathbf{Z}_A$, we remove the known interactions by setting the entries of $\mathbf{Z}_A$ corresponding to nonzero entries in $\mathbf{A}$ to zeros and sort the remaining scores in descending order. The drugs-target pairs with highest scores are the most likely unknown interacting pairs. The full process of the proposed method is illustrated in Fig. \ref{fig_framework}. The algorithm 1 illustrates the detailed procedure of the proposed method LMP.

\begin{table}
\caption{The illustration of the LMP algorithm}
\begin{tabular}{p{2.1cm} p{1.1cm} p{12.0cm}}
\hline \\[-5.5ex]
\multicolumn{3}{l}{\textbf{Algorithm 2}: The algorithm of the proposed method} \\
\hline \\[-5.5ex]
\multicolumn{3}{l}{\textbf{Input:} Given an adjacency matrix $\mathbf{A}$} \\
& 1. & compute the low-rank similarity matrix $\mathbf{X}^*$ and  \\
& &  sparse noise $\mathbf{E}^*$ of Eq. (\ref{eq_unconstraint_lrr}) by using Algorithm 1\\
& 2. & compute the similarity matrices $\mathbf{Z}_{AD}$ and $\mathbf{Z}_{AT}$ by Eq. (\ref{eq_similarity}) \\
& 3. & combine the two similarity matrices as in Eq. (\ref{eq_similarity_com})\\
& 4. & sort the scores in $\mathbf{Z}_A$ in descending order\\
\multicolumn{3}{l}{\textbf{Output:} The highest scores are the most potential interactions} \\[0.3ex]
\hline
\end{tabular}
\label{table_LMP}
\end{table}

\subsection{Working with Heterogeneous Data}
Using only interaction dataset and ignoring the extra characteristic information of the drugs and targets is throwing away the important information. In this subsection, we show how the proposed method is capable of utilizing this characteristic information. Based on the hypothesis that similar drugs interact with similar targets and vice versus, we can utilize the two kinds of characteristic information, namely drug similarity and target similarity to infer the potential interactions.

Drug similarity $\mathbf{S}_D$ was computed by using SIMCOMP \cite{hattori2003development} from the chemical structures of drugs which are obtained from KEGG LIGAND \cite{kanehisa2006genomics}. On the other hand, target similarity $\mathbf{S}_T$ is computed by using a normalized Smith-Waterman score \cite{smith1981identification} from GEGG GENES \cite{kanehisa2006genomics}. These two datasets are available online \cite{yamanishi2008prediction}. Directly using these two similarity datasets makes the proposed method perform unsatisfactorily since there exist some noise inside these two datasets. Therefore, we compute the new similarity matrices which are low-rank from these two similarity matrices by the proposed method, then projecting the adjacency matrix onto these lower-dimensional spaces. There are two main properties of this low-rank matrix learning from the characteristic information. Firstly, as discussed above the noise are subtracted. Moreover, the interaction information is projected on the lower-dimensional feature space which is more informative.

The low-rank similarity matrices of the characteristic information $\mathbf{S}_D$ and $\mathbf{S}_T$ can be computed as shown in Eq. (\ref{eq_lrr}) by replacing the adjacency matrix $\mathbf{A}$ with $\mathbf{S}_D$ and $\mathbf{S}_T$, respectively. After obtaining the low-rank similarity matrices of the drug and target denoted as $\mathbf{X}^*_{SD}$ and $\mathbf{X}^*_{ST}$, we project the adjacency matrix $\mathbf{A}$ onto them as shown in Eq. (\ref{eq_similarity}) and we call them the score matrices denoted as $\mathbf{Z}_D$ and $\mathbf{Z}_T$, respectively. Finally, we combine all these three score matrices which are $\mathbf{Z}_A$, $\mathbf{Z}_D$ and $\mathbf{Z}_T$ as
\begin{equation}
    \mathbf{Z}_{ADT} = \gamma_1\mathbf{Z}_{A} +\gamma_2 \mathbf{Z}^T_{D} + \gamma_3\mathbf{Z}_{T},
\end{equation}
where $\gamma_1$, $\gamma_2$ and $\gamma_3$ are the weighting parameters and are set to 0.5, 0.25 and 0.25, respectively. Since the known interactions are experimentally verified, the similarity matrix obtained from this information plays more important role than the other two similarity matrices. The algorithm of the proposed method is illustrated in table \ref{table_LMP}.

\subsection{Predicting the New Drugs and Targets}
In case that there are new drugs or targets which do not have any known interactions at all, the proposed method can also be simply extended to predict their interactions. However, we need to use the characteristic information about the new drugs or new targets. On one hand, once we are given a drug with characteristic information, we wish to predict which target proteins that this drug interactions with. On the other hand, we aim at predicting the new drugs based on the known protein target by using the characteristic information of the target proteins. Consider predicting new targets, first of all, one needs to compute the low-rank similarity matrix of the given drug with others based on their characteristic information, i.e., computing $\mathbf{X}^*_{SD}$. With the assumption that similar drugs interact with similar targets, we can predict the potential interactions based on this similarity matrix, i.e., projecting $\mathbf{A}$ onto $\mathbf{X}^*_{SD}$. Similarly, when a new target is given with its biological information, one can compute the low-rank similarity matrix, i.e., $\mathbf{X}^*_{ST}$, of that protein with the others. The potentially interacted drugs with this protein are those that interact with proteins that are most similar to the given protein.

\begin{table*}[ht!]
\centering
\small{
\caption{ The interaction predicted results measured by AUC and AUPR by 5$\times$10-fold cross validation for LMP and the benchmark methods. In the first parts only the known interactions are utilized, while in the second parts characteristic information is also employed. The best performed results are emphasized in bold.}
    \begin{tabular}{p{2.0cm} p{0.7cm}  p{0.9cm}  p{0.7cm} p{0.9cm}   p{0.7cm} p{0.9cm} p{0.7cm}  p{0.9cm} p{0.7cm} p{1.2cm} p{0.7cm} p{1cm}}
      \hline \\ [-5.5ex]
             Dataset & \multicolumn{2}{c}{MATADOR}  &  \multicolumn{2}{c}{NR} & \multicolumn{2}{c}{GPCR}& \multicolumn{2}{c}{ion channel} & \multicolumn{2}{c}{enzyme} & \multicolumn{2}{c}{Average} \\ [0.5ex]
        \hline \\ [-5.5ex]
            Metric  & AUC   & AUPR &  AUC   & AUPR  & AUC   & AUPR &  AUC   & AUPR & AUC   & AUPR & AUC   & AUPR  \\ [0.5ex]
        \hline \\ [-5.5ex]
            CN	                        & 0.930 & 0.603 & 0.688 & 0.249 & 0.826 & 0.491 & 0.915 & 0.724 & 0.910 & 0.678 & 0.854 & 0.549 \\
            Katz                        & 0.894 & 0.393 & 0.679 & 0.263 & 0.800 & 0.479 & 0.893 & 0.707 & 0.869 & 0.644 & 0.827 & 0.497  \\
            Jaccard                     & 0.933 & 0.612 & 0.686 & 0.249 & 0.828 & 0.531 & 0.920 & 0.680 & 0.911 & 0.704 & 0.856 & 0.555 \\
            CJC                         & 0.930 & 0.609 & 0.693 & 0.265 & 0.828 & 0.531 & 0.914 & 0.731 & 0.910 & 0.673 & 0.855 & 0.562  \\
            CRA                         & 0.937 & 0.714 & 0.696 & \textbf{0.289} & 0.833 & 0.565 & 0.923	& 0.737 & \textbf{0.912} & 0.752 & 0.860 & 0.612 \\
            LMP-$\mathbf{Z}_A$          & \textbf{0.946} & \textbf{0.796} & \textbf{0.702} & 0.276 & \textbf{0.853} & \textbf{0.601} & \textbf{0.941} & \textbf{0.846} & 0.900 & \textbf{0.766} & \textbf{0.868} & \textbf{0.657} \\ [0.5ex] 

        \hline \\ [-4.8ex]
            BLM                        & -- & -- & 0.694 & 0.204 & 0.884 & 0.464 & 0.918 & 0.591 & 0.928 & 0.496 & 0.856 & 0.439 \\
            LapRLS                     & -- & -- & 0.855 & 0.539 & 0.941 & 0.640 & 0.969 & 0.804 & 0.962 & 0.826 & 0.932 & 0.702 \\
            NetLapRLS                  & -- & -- & 0.859 & \textbf{0.563} & 0.946 & 0.703 & 0.977 & 0.898 & 0.968 & \textbf{0.874} & 0.938 & \textbf{0.760} \\
            LMP-$\mathbf{Z}_{ADT}$     & -- & -- & \textbf{0.863} & 0.513 & \textbf{0.950} & \textbf{0.706} & \textbf{0.979} & \textbf{0.900} & \textbf{0.973} & \textbf{0.875} & \textbf{0.941} & 0.747 \\ [0.5ex]
        \hline
      \end{tabular}
      \label{table_pre}
}
\end{table*}

\subsection{Evaluation and Experimental Settings}
We adopt a cross validation technique and two popular metrics to test the proposed method as well as previous benchmarks. We apply the 10-fold cross validation \cite{ding2013similarity,lu2017link}, which divides the total known interactions between the chemicals and proteins into 10 sets with approximately the same size, and then utilize 9 sets as training data and keep the remaining set as testing data. We repeat it for ten times where each set has one chance to be the testing set. In the simulation, we independently run the 10-fold cross validation for five times and report average values accordingly.

We consider the two popular metrics, including the area under the receiver operating characteristic (ROC) curve (AUC) and the area under precision and recall curve (AUPR), to evaluate the performances of the proposed method and the benchmarks. ROC is the diagnostic ability of a binary classifier with regarding to different thresholds \cite{hanley1982meaning}, while AUC curve displays true positive rate (sensitivity) versus false positive rate (1-specificity) at different values of thresholds. The sensitivity is the percentage of the test samples with ranks higher than a given threshold, whereas, specificity is the percentage the test samples that fall below the threshold. When there are many fewer positive elements in the testing data comparing to the total number in testing data, AUC may give overoptimistic results of the algorithms \cite{bleakley2009supervised,davis2006relationship,lu2012recommender,lobo2008auc}. Therefore, utilizing only AUC may mislead our conclusion. In such case, AUPR can give better evaluation, especially in biological significance.

The simulations are conducted within three manners, e.g., drug-target pairs, new drugs, and new targets. In the first manner, we divide the total known interactions into 10-folds with approximately the same size. On the other hand, in the second manner we divide the total drugs into 10-folds. For the last manner, all the targets are divided into 10-folds. In each simulation, we use 9 sets as training data and keep the remaining as testing data.

\subsection{Baseline Methods}
First, we compare LMP with the similarity-based methods, e.g., common neighbor index (CN), Katz index and Jaccard index, Cannistraci resource allocation (CRA) \cite{daminelli2015common} and Cannistraci Jaccard index (CJC) \cite{daminelli2015common}. In this comparison, we use only the known interactions of drug and targets. The methods that utilize both interaction information and characteristic information normally outperform the methods that utilize only the known interactions. However, they cannot work with the dataset that do not have characteristic information such as MATADOR. We further compare LMP with bipartite local learning model (BLM) \cite{bleakley2009supervised}, Laplacian regularized least squares (LapRLS) and Net Laplacian regularized least squares (NetLapRLS) \cite{xia2010semi} which utilizes both the known interaction information and characteristic information about the drugs and targets. The first group of methods can only predict the interactions between drugs and targets, meanwhile the second group can predict the new drugs given targets or predict the new targets given drugs by using their characteristic information.

\begin{table*}[t!]
\centering
\caption{ The drug and target predicted results measured by AUC and AUPR by 5$\times$10-fold cross validation for LMP and the benchmark methods. The best performed results are emphasized in bold.}
    \begin{tabular}{p{2.2cm} p{0.7cm} p{0.9cm} p{0.7cm} p{0.9cm} p{0.7cm} p{0.9cm} p{0.7cm} p{1.2cm} p{0.7cm} p{1cm}}
      \hline \\ [-5.5ex]
             Dataset & \multicolumn{2}{c}{nuclear receptors} & \multicolumn{2}{c}{GPCR}& \multicolumn{2}{c}{ion channel} & \multicolumn{2}{c}{enzyme} & \multicolumn{2}{c}{Average} \\ [0.5ex]
        \hline \\ [-5.5ex]
            Metric  &  AUC   & AUPR & AUC   & AUPR &  AUC   & AUPR  & AUC   & AUPR & AUC   & AUPR \\ [0.5ex]
        \hline \\ [-5.5ex]
            BLM                         & 0.693 & 0.194 & 0.829 & 0.210 & 0.770 & 0.167 & 0.781 & 0.092 & 0.768 & 0.166 \\
            LapRLS                      & 0.820 & \textbf{0.482} & 0.845 & 0.397 & \textbf{0.796} & \textbf{0.366} & {0.800} & 0.368 & 0.815 & \textbf{0.403}  \\
            NetLapRLS                   & 0.819 & 0.418 & 0.834 & 0.397 & 0.783 & 0.343 & 0.791 & 0.298 & 0.807 & 0.364 \\
            LMP-$\mathbf{Z}_{T}$	    & \textbf{0.831} & 0.384 & \textbf{0.854} & \textbf{0.399} & 0.778 & 0.353 & \textbf{0.824} & \textbf{0.392}  & \textbf{0.822} & 0.382 \\ [0.5ex]

             \hline \\ [-5.5ex]
            BLM                         & 0.458 & 0.325 & 0.627 & 0.367 & 0.881 & 0.641 & 0.843 & 0.611 & 0.702 & 0.486 \\
            LapRLS                      & 0.563 & 0.432 & 0.788 & 0.508 & 0.920 & 0.778 & 0.914 & 0.792 & 0.796 & \textbf{0.628} \\
            NetLapRLS                   & 0.561 & \textbf{0.433} & 0.787 & 0.503 & 0.916 & 0.762 & 0.909 & 0.787 & 0.793 & 0.621 \\
            LMP-$\mathbf{Z}_{D}$        & \textbf{0.688} & 0.293 & \textbf{0.847} & \textbf{0.583} & \textbf{0.938} & \textbf{0.790} & \textbf{0.928} & \textbf{0.803} & \textbf{0.850} & 0.617 \\ [0.5ex]
        \hline
      \end{tabular}
      \label{table_preTarget}
\end{table*}

\begin{figure*}[t!]
\centering
	\includegraphics[width=0.95\textwidth]{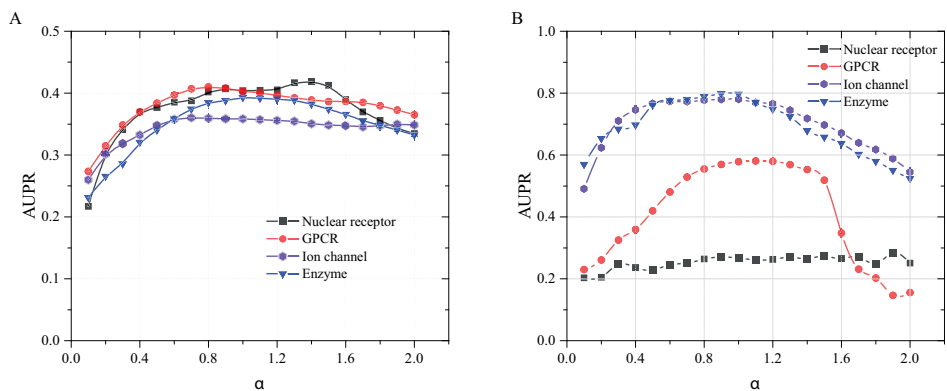}
    \caption{The AUPR of the proposed method on the four datasets. Panel A illustrates the predicted results on the new drugs, while panel B displays the predicted results on the new targets.}
\label{fig_auprAlpha}
\end{figure*}

\subsection{Parameter Settings}
In the proposed method, there is a free parameter $\alpha$ that balances the weights of low-rank matrix and sparse matrix as shown in Eq. (\ref{eq_lrr}). When $\alpha$ is set too large, the sparse norm will compress most of the entries of matrix $\mathbf{E}$ to zeros, while if $\alpha$ is very small, most of the entries of $\mathbf{E}$ will be small but not zeros. In this work, we obtain the optimal value of the parameter $\alpha$ by manually and empirically tuning it and check the accuracy according to each value of $\alpha$. Predicting the interactions based on only the known interaction, we perform the grid search for $\alpha$ in which one dimension is corresponding to drug similarity and another one is corresponding to target similarity. Based on the empirical simulation, $\alpha$ for predicting the interactions based on only known interactions falls between [0.1,0.25]. When characteristic information is embedded, one needs to tune $\alpha$ according to this information, e.g., the parameter $\alpha$ falls between [0.1,2]. Moreover, there is only one $\alpha$ in each case, eg., predicting the new drugs or targets. We visualize the sensitivity of $\alpha$ corresponding to the predicted results, i.e., AUPR, on the new drugs and targets, in Fig. \ref{fig_auprAlpha}.

\section{Results}

We report the performances of the proposed method and the others that use only the known interaction information in the first part of table \ref{table_pre}, meanwhile for the performances of the methods that use together interaction information and characteristic information are shown in the second part of the same table. Consider the first group of the methods. LMP outperforms the other methods on three datasets including MATADOR, GPCR, and ion channel in terms of AUC and AUPR. In term of AUPR, LMP outperforms the others in enzyme,  while LMP outperforms only Katz index in enzyme in term of AUC. It is worth noting that for the small matrix or network such as nuclear receptor, the predicted results from all the methods are not stable and they are approximately the same.

When the characteristic information of the drugs and targets are employed, the performances of the LMP remarkably improved as shown in the second part of table \ref{table_pre}. Since the characteristic information of the drugs and targets in MATADOR is not available, we either cannot obtain the results of this dataset. LMP outperforms the others on nuclear receptor and GPCR in term of AUC, while NetLapRLS is the best in term of AUPR on nuclear receptor. LMP perform better than the others measured by AUPR in GPCR and ion channel. LMP outperforms the others in term of AUC, while NetLapRLS produces the highest AUPR on enzyme.

The predicted results on the new drugs and target proteins are illustrated in table \ref{table_preTarget} in the first and second part, respectively. All the methods produce competitive results in predicting the new drugs, while LMP outperform the others in term of AUC and AUPR on GPCR, ion channel and enzyme in predicting new targets.

\section{Conclusion and discussion}

In this work, we have proposed a matrix-based method, namely low-rank matrix projection (LMP), to solve the DTI prediction problem. It has been shown that LMP overcomes the drawbacks of the machine learning and similarity-based methods. On one hand, LMP can work on datasets that have only known interaction information between the drugs and targets such as MATADOR. On the other hand, LMP can integrate the information about the characteristics of the drugs and targets to improve the predicted results. Moreover, the proposed method can also effectively deal with the new drugs and targets that do not have any known interaction at all by utilizing only some characteristic information of the drugs or targets. 

In LMP, the low-rank matrix plays a very important role in making the data homogenous, meanwhile the sparse matrix captures the noise or outliers in the data. By decomposing the original data into a clean (a linear combination of low-rank matrix and the adjacency matrix) and noise (sparse matrix) parts, we can obtain a clean data to predict the interactions between drugs and target proteins. The disadvantage of LMP is that we need to empirically tune $\alpha$ and check the accuracy corresponding to each value of $\alpha$. Until now, designing the effective method to estimate the optimal value of this parameter is still an open question. Moreover, LMP may not perform well with small matrix, e.g., nuclear receptor, since the information is very limited for matrix decomposition.

LMP is an alternative, effective and efficient \textit{in silico} tool for predicting the drug-target interactions. It can help drug development and drug reposition. In this paper, LMP aims at learning the low-rank similarity matrices from the known drug-target interaction information and similarity matrices of the drugs and targets. We believe that the proposed method can also be applied to learn other high-dimensional biological data such as drug compound, chemical structures, and so on, and we leave these problems to the future work.

\end{document}